\pdfoutput=1
\documentclass{article}
\PassOptionsToPackage{numbers,compress}{natbib}

\usepackage[preprint]{neurips_2026}

\usepackage[utf8]{inputenc}
\usepackage[T1]{fontenc}
\usepackage{hyperref}
\usepackage{url}
\usepackage{booktabs}
\usepackage{amsmath}
\usepackage{amssymb}
\usepackage{amsfonts}
\usepackage{nicefrac}
\usepackage{microtype}
\usepackage{xcolor}
\usepackage{pifont}
\usepackage{tikz}
\usetikzlibrary{positioning,arrows.meta,shapes.geometric,fit,backgrounds,
                 decorations.pathreplacing,calc}
\usepackage{pgfplots}
\pgfplotsset{compat=1.18}
\usepackage{capt-of}

\newsavebox{\tikzfitbox}
\let\origtikzpicture\tikzpicture
\let\origendtikzpicture\endtikzpicture
\renewenvironment{tikzpicture}[1][]{%
  \begin{lrbox}{\tikzfitbox}%
  \origtikzpicture[#1]%
}{%
  \origendtikzpicture%
  \end{lrbox}%
  \ifdim\wd\tikzfitbox>\linewidth
    \resizebox{\linewidth}{!}{\usebox{\tikzfitbox}}%
  \else
    \usebox{\tikzfitbox}%
  \fi
}
\usepackage{array}
\usepackage{multirow}
\usepackage{enumitem}

\newcommand{\R}{\mathbb{R}}
\newcommand{\softmax}{\operatorname{softmax}}
\newcommand{\mem}{\mathrm{Mem}}
\newcommand{\freeze}{\mathrm{frozen}}
\newcommand{\topk}{\operatorname{top\text{-}k}}

\title{Trained Persistent Memory for Frozen Encoder--Decoder LLMs:\\
Six Architectural Methods}

\author{%
  Hong Jeong \\
  Inha University in Tashkent, Uzbekistan \\
  \texttt{hjeong@postech.ac.kr}
}

\begin{document}
\maketitle

\begin{abstract}
Frozen encoder--decoder language models are stateless: the latent
representation is discarded after every forward pass, so no information
persists across sessions.  This paper presents a \textbf{proof-of-concept
pilot study} showing that persistent memory in the \emph{continuous
latent space} of a frozen LLM is feasible---even under severe resource
constraints (a single frozen Flan-T5-XL backbone, small trainable
adapters, a single dataset).  We implement six architectural methods
spanning three injection points and four write mechanisms; unlike
text-level memory systems, every write and read is a differentiable
operation on dense vectors.  After training only the adapter, the memory
bank continues to accumulate at inference time without gradients,
enabling \emph{conversational learning}.  Under a forgetting-curve
evaluation on LoCoMo at two capacity scales (1$\times$ and
10$\times$), the stateless baseline scores exactly zero; at
10$\times$ all six trained adapters produce positive memory-recall
curves; at 1$\times$ three methods collapse, revealing capacity as a
critical design parameter.  Because the memory bank is a compact
numerical array, it can be scaled to arbitrarily large capacity
without altering the backbone.  We argue that full end-to-end training
with larger models, larger data, and orders-of-magnitude larger memory
will yield substantially stronger results; this pilot study establishes
the feasibility baseline and design-space taxonomy that such
efforts require.
\end{abstract}

\section{Introduction}
\label{sec:intro}

Consider a frozen encoder--decoder model such as Flan-T5 built on a
T5-style backbone~\citep{raffel2020t5}.  The forward pass is:
\begin{equation}
  Z_t = E_{\freeze}(x_t), \qquad \hat{y}_t = D_{\freeze}(Z_t),
  \label{eq:baseline}
\end{equation}
where $E_{\freeze}$ and $D_{\freeze}$ are fixed pre-trained weights,
$x_t$ is the input at turn~$t$, $Z_t \in \R^{n \times d}$ is the encoder
output, and $\hat{y}_t$ is the generated text.  This system is
\textbf{stateless}: $Z_t$ is discarded after each forward pass and the
model has no recollection of previous turns.  If a user says ``I like
reading'' in session~1 and asks ``What do I like?'' in session~3, the
model cannot answer---there is no state that survives across sessions.
This \emph{inter-session memory} problem is the concrete target of
persistent memory.

Existing long-term memory solutions such as MemGPT and MemoryBank
operate at the \emph{text level}: they store, summarize, and retrieve
natural-language passages through an external database.  This paper
works at a fundamentally different level---the \textbf{latent space}
of the frozen model.  The memory bank $P_t \in \R^{n_P \times d}$
holds continuous encoder representations, not strings, so writing and
reading are differentiable operations embedded inside the forward pass
rather than pre- or post-processing steps around it.

A persistent bank built directly from frozen encoder outputs is not
enough.  The decoder's cross-attention was pre-trained to read current
encoder states, not arbitrarily accumulated cache states, so a naive
memory path tends to dilute attention rather than sharpen retrieval as
history grows.  Recent attention-coupled latent-memory work shows that
learned structure inside the memory pathway can instead induce functional
specialization and controlled routing~\citep{jeong2026lateral,jeong2026brain}.
For frozen encoder--decoder LLMs, this suggests that a small trainable
adapter is the minimal mechanism needed to write memory in a form the
frozen decoder can use.

This paper takes the necessary next step: we \textbf{allow training} for
a small memory adapter~$\theta_{\mem}$ while keeping both encoder and
decoder frozen.  We augment the stateless system with a
\textbf{persistent memory bank}
$P_t \in \R^{n_P \times d}$ that persists across turns and sessions:
\begin{equation}
  Z_t = E_{\freeze}(x_t), \qquad
  P_t = \mathrm{Write}(P_{t-1}, Z_t), \qquad
  \hat{y}_t = D_{\freeze}\!\bigl(\mathrm{Read}(Z_t, P_{t-1})\bigr).
  \label{eq:augmented}
\end{equation}
The \texttt{Write} operation updates~$P$ from the current latent; the
\texttt{Read} operation injects historical context from~$P$ into the decoder.
The learned parameters~$\theta_{\mem}$ enable the adapter to \emph{learn
how to write memory in a format that the frozen decoder's existing
cross-attention can discriminate}---the capability that zero-training
methods provably lack.

This paper is deliberately designed as a \textbf{low-budget pilot study}:
we use a single frozen backbone, a single evaluation dataset, and
minimal adapter parameters.  The goal is not to achieve
state-of-the-art recall but to \emph{demonstrate feasibility}---that
persistent latent-space memory can be installed in an existing frozen
LLM with inexpensive adapters, and that the resulting system
exhibits non-trivial, capacity-dependent memory behaviour.  Full
end-to-end training with unfrozen, larger-scale LLMs, bigger
datasets, and memory banks orders of magnitude larger than ours lies
beyond the scope of this pilot but is the natural industrial-scale
follow-up that our results motivate.

Empirically, even under these constrained conditions the six designs
separate clearly.  At 10$\times$ capacity all trained adapters rise
above the zero baseline on the forgetting curve; at 1$\times$ three
methods collapse.  M.2~XAttn and M.6~Slot dominate at low capacity,
while M.4~Hebbian leads at high capacity---revealing memory-bank size
as a critical design parameter.

After training, $\theta_{\mem}$ is frozen but $P_t$ continues to
accumulate at inference time without gradients.  We call this
\textbf{conversational learning}: each new session enriches~$P$, so a
fact stated in session~1 (``I am John Doe'') can be recalled in session~10
(``Who am I?'') without re-stating it and without million-token context
windows---the relevant inter-session history is compressed into~$P$.

The analogy to human cognition is deliberate.  Human brains accumulate
knowledge through complementary memory systems~\citep{tulving1972episodic}:
\emph{episodic} memory records specific events (``Alice mentioned a trip
to Paris''), \emph{semantic} memory distills general facts (``Alice likes
travel''), \emph{procedural} memory encodes how-to skills, and
\emph{working memory} holds the active conversational context.  All of
these are linked through \emph{associative} retrieval: a cue in working
memory can trigger recall from any long-term store.  The persistent memory
bank~$P$ in our framework plays an analogous role: the write rule
determines what is stored (episodic vs.\ semantic), the read rule
implements associative retrieval, and the capacity of~$P$ constrains how
much can be retained---mirroring the interplay of encoding, consolidation,
and retrieval in biological memory.  A successful method must demonstrate
not merely storage but \emph{remembering} (retrieving the right fact),
\emph{generalisation} (answering questions phrased differently from the
original statement), and \emph{abstraction} (combining multiple facts
into a coherent response)---the hallmarks of genuine learning from
experience.

The six methods differ along three orthogonal design dimensions:
(i)~\emph{where} $P$ enters the forward pass (before the encoder,
between encoder and decoder, or inside the decoder),
(ii)~\emph{how} $P$ is written (attention-coupled update, Hebbian outer
product, gated cross-attention, or sparse slot addressing), and
(iii)~\emph{how many parameters are added} (all modest relative to the
3B-parameter backbone).
A crucial constraint across all methods is that the frozen decoder is
calibrated exclusively for encoder outputs.  Let $\mathcal{M}_{E}$ denote
the set of representations the encoder can produce; any method that
replaces~$Z$ with $H \notin \mathcal{M}_{E}$ will degrade the decoder.
All six methods preserve the frozen encoder--decoder route through~$Z_t$
and ensure that memory influence enters through controlled, learnable
pathways.

Our core evaluation is simple: \textbf{how much does each method
remember?}  Every condition---baseline and all six memory methods---sees
only the current turn~$x_t$; no method receives the conversation history.
The baseline is deliberately short-sighted and retains nothing.  Each
memory method must accumulate facts into~$P$ through its write rule and
retrieve them through its read path.  By probing the same factual
questions across increasing evidence lag, we measure a
\emph{forgetting curve}: the fraction of available headroom that the
method's persistent state fills, normalised so that 100\% means
memory brings F1 to the gold standard and 0\% means memory adds
nothing.  The stateless baseline has no persistent state and is
therefore identically zero; stronger methods start higher at short lag
and decay more slowly as the evidence recedes into the past.

A secondary question is whether the frozen decoder's cross-attention
queries---trained only on encoder outputs---have enough
\textbf{representational slack} to attend usefully to memory entries
projected by the trained adapter.  If not, even perfectly trained
$\theta_{\mem}$ will not help, implying that the frozen decoder itself is
the bottleneck.

\medskip\noindent\textbf{Contributions.}
\begin{enumerate}[leftmargin=*,nosep]
\item \textbf{Latent-space persistent memory.}
  We formulate the problem of adding persistent memory that lives entirely
  in the continuous latent space of a frozen encoder--decoder LLM.
  Unlike text-level memory systems (MemGPT, MemoryBank) that store and
  retrieve natural-language strings outside the model, our memory bank
  $P_t \in \R^{n_P \times d}$ holds dense encoder representations; every
  write and read is a differentiable operation inside the forward pass.

\item \textbf{Six architectural methods.}
  We design, implement, and release six trained memory adapters that span
  three injection points (before the encoder, between encoder and decoder,
  inside the decoder) and four write mechanisms (attention-coupled update,
  Hebbian outer product, gated cross-attention, sparse slot addressing).
  All methods keep every encoder and decoder weight frozen and add only a
  small set of learnable parameters~$\theta_{\mem}$.

\item \textbf{Headroom-normalised forgetting-curve evaluation.}
  We introduce an evaluation protocol that measures the fraction of
  available answer-quality headroom filled by a method's persistent
  state, as a function of evidence lag.  The metric is normalised to a
  0--100\% scale (100\%~=~perfect recall, 0\%~=~no memory
  contribution), giving intuitive and comparable scores.  The stateless
  baseline is identically zero by construction.

\item \textbf{Empirical findings.}
  Under this protocol on LoCoMo, we test at two capacity scales
  (1$\times$ and 10$\times$).  At 10$\times$ all six trained adapters
  produce positive memory-recall curves; at 1$\times$ three methods
  collapse, revealing capacity as a critical design parameter.
  M.2~XAttn and M.6~Slot dominate at low capacity; M.4~Hebbian leads
  at high capacity.  Knowledge accumulation curves confirm that the best
  methods steadily accumulate facts over 30 sessions.
\end{enumerate}

\section{Related Work}
\label{sec:related}

Persistent memory is adjacent to, but distinct from, several existing lines of
work.  Application-level long-term memory systems such as
MemGPT~\citep{packer2024memgpt} and
MemoryBank~\citep{zhong2024memorybank} demonstrate that explicit memory can
improve LLM behaviour over extended interactions, but they operate at the
\emph{text level}: facts are stored as natural-language strings, retrieval is
a search over those strings, and the language model itself is unchanged.
LoCoMo~\citep{maharana2024locomo} provides a
public benchmark targeted specifically at very long-term conversational memory
and motivates the multi-session evaluation setting adopted here.

The present paper operates at a fundamentally different level---the
\textbf{latent space} of the frozen model.  Our persistent memory bank stores
continuous encoder representations, not text; reading and writing are
differentiable operations inside the forward pass rather than pre- or
post-processing steps.  Rather than proposing a single end-to-end memory
agent, we define a taxonomy of six architectural alternatives for
latent-space persistent memory, formalize their read and write paths, and
specify a public-dataset protocol for comparing them against a stateless
baseline under a shared released backbone.

\paragraph{Parameter-efficient adaptation.}
Several of our six methods adapt ideas originally proposed for
parameter-efficient fine-tuning or memory-augmented architectures.
Prefix tuning~\citep{li2021prefix} prepends learnable soft tokens to
the input; M.1 extends this idea to a persistent memory bank.
Flamingo~\citep{alayrac2022flamingo} inserts gated cross-attention
layers into a frozen decoder for visual grounding; M.2 and M.5 adopt the
same parallel-branch topology for memory injection.
Memorizing Transformers~\citep{wu2022memorizing} extend the decoder KV
cache with retrieved past representations; M.3 follows the same KV
concatenation principle.  Linear Transformers and fast weight
programmers~\citep{schlag2021linear} accumulate an outer-product
associative matrix updated at every step; M.4 uses the same Hebbian
write rule.  Neural Turing Machines~\citep{graves2014neural} maintain
addressable memory slots with content-based sparse writes; M.6 inherits
this slot-addressing mechanism.  Crucially, none of these prior methods were designed for \emph{persistent
latent-space} memory that accumulates across sessions inside a frozen
encoder--decoder model.  Our contribution is not the individual read or
write primitive but the controlled comparison---under a single frozen
backbone and a common forgetting-curve evaluation---of how these
primitives perform as latent-space persistent memory for inter-session
recall.

\paragraph{Attention-coupled latent memory.}
Recent arXiv work on attention-coupled latent memory explores richer
structured bank dynamics than the minimal adapters studied here.
Inhibitory cross-talk across paired banks can drive functional
lateralization~\citep{jeong2026lateral}, while a miniature brain-transformer
architecture adds thalamic gating, amygdaloid salience, hippocampal
lateralization, and prefrontal working memory to shape routing and
consolidation~\citep{jeong2026brain}.  The present paper is narrower: it
compares six simpler trainable memory adapters under a fixed frozen
encoder--decoder backbone and a common forgetting-curve evaluation.

\paragraph{Cognitive memory systems.}
Our taxonomy is informed by the cognitive science of human memory.
Tulving's distinction between episodic and semantic
memory~\citep{tulving1972episodic} maps onto our design choices:
methods with high-capacity slot banks (M.6) store episode-like snapshots,
while Hebbian associative memory (M.4) naturally distils
co-occurrence statistics akin to semantic memory.  The content-gated
method~(M.5) resembles working-memory gating, selectively admitting
relevant information.  Modern complementary learning systems
theory~\citep{mcclelland1995cls, kumaran2016cls} argues that fast
episodic binding and slow semantic consolidation are both necessary;
our experiment measures whether any single-mechanism method suffices
or whether the task demands a composite architecture.

\section{Problem Setting and Notation}
\label{sec:baseline}

Figure~\ref{fig:baseline} shows the stateless frozen encoder--decoder pipeline
used as the control architecture throughout the paper.

\begin{center}
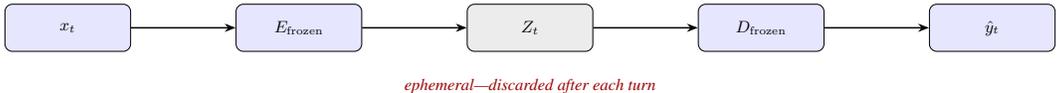

\begin{tikzpicture}[
  node distance=1.0cm and 2.0cm,
  block/.style={draw, rounded corners, minimum width=2.4cm,
                minimum height=0.9cm, align=center, font=\small},
  frozen/.style={block, fill=blue!10},
  latent/.style={block, fill=gray!15},
  arr/.style={-{Stealth[length=6pt]}, thick},
]
  \node[frozen] (x) {$x_t$};
  \node[frozen, right=of x] (enc) {$E_{\freeze}$};
  \node[latent, right=of enc] (Z) {$Z_t$};
  \node[frozen, right=of Z] (dec) {$D_{\freeze}$};
  \node[frozen, right=of dec] (y) {$\hat{y}_t$};

  \draw[arr] (x) -- (enc);
  \draw[arr] (enc) -- (Z);
  \draw[arr] (Z) -- (dec);
  \draw[arr] (dec) -- (y);

  \node[below=0.4cm of Z, font=\small\itshape, text=red!60!black]
    {ephemeral---discarded after each turn};
\end{tikzpicture}

\captionof{figure}{Frozen encoder--decoder baseline used as the stateless
control. The latent representation is consumed within the current turn and then
discarded.}
\label{fig:baseline}
\end{center}

\noindent
The encoder maps $x_t$ to $Z_t \in \R^{n \times d}$ (sequence length~$n$,
hidden dimension~$d$).  The decoder uses cross-attention to read~$Z_t$ at every
layer.  Both $E_{\freeze}$ and $D_{\freeze}$ are frozen; only the persistent
memory $P$ and (optionally) a small set of adapter
parameters $\theta_{\mem}$ are modified.

\section{Trained Alternatives}
\label{sec:trained}

These methods introduce a small set of learnable parameters
$\theta_{\mem}$ trained via backpropagation; at inference time (Type~2),
$P$ continues to accumulate without gradients.

Every method uses \emph{content-based} addressing---deciding what to store
or retrieve based on the semantic content of the current turn rather than
fixed positions.
Table~\ref{tab:readwrite} gives the complete read and write operations for
all six methods; the baseline (M.0) is included for reference.

\begin{table}[!hbtp]
\centering
\caption{Read and write operations for each method.
$A = \softmax(Z_t W_Q (P W_K)^\top / \sqrt{d})$.
\emph{Delegated} read means the frozen decoder cross-attention selects from
the concatenated KV; \emph{explicit} read means the adapter performs its own
retrieval before passing to the decoder.}
\label{tab:readwrite}
\small
\renewcommand{\arraystretch}{1.25}
\resizebox{\textwidth}{!}{%
\begin{tabular}{@{}c l l l l@{}}
\toprule
& \textbf{Method}
  & \textbf{Write rule} ($P_t \leftarrow$)
  & \textbf{Read rule} (inject into decoder)
  & \textbf{Type} \\
\midrule
0 & Baseline
  & ---
  & $D(Z_t)$
  & --- \\
1 & Prefix
  & $\gamma P + A^\top V$,\; $V{=}Z_t W_V$
  & $[Z_t;\; P\,W_P] \to D$  (extra KV)
  & Delegated \\
2 & XAttn
  & $\gamma P + A^\top V$,\; $V{=}Z_t W_V$
  & $\softmax(s\,W_Q^m (P\,W_K^m)^\top\!/\!\sqrt{d})\,(P\,W_V^m)$
  & Explicit \\
3 & KV Extension
  & $\gamma P + A^\top V$,\; $V{=}Z_t W_V$
  & $K{=}[K_Z;\,P W_{K,m}],\; V{=}[V_Z;\,P W_{V,m}]$
  & Delegated \\
4 & Hebbian
  & $\gamma M + (Z_t W_{K,H})^\top (Z_t W_{V,H})$
  & $(Z_t W_{Q,H})\,M \to W_{\mem} \to D$ (extra KV)
  & Explicit \\
5 & Gated
  & $\gamma P + A^\top V$,\; $V{=}Z_t W_V$
  & $g_t \odot \mathrm{XAttn}(s, P)$,\; $g_t{=}\sigma(W_g[s;c]+b_g)$
  & Explicit \\
6 & Slot
  & top-$k$: $\gamma P[s] + (1{-}\gamma)\,\bar{z}_t W_u$
  & $K{=}[K_Z;\,P W_{K,m}],\; V{=}[V_Z;\,P W_{V,m}]$
  & Delegated \\
\bottomrule
\end{tabular}}%
\end{table}

\noindent
\emph{Delegated-read} methods (M.1, M.3, M.6) project all of~$P$ into
the decoder's KV cache and let the frozen cross-attention select relevant
entries.  \emph{Explicit-read} methods (M.2, M.4, M.5) perform their own
content-based retrieval before passing the result to the decoder.  This
distinction has implications for trainability: explicit-read methods
introduce more parameters but give the adapter direct control over
retrieval; delegated-read methods rely on the decoder's pre-trained
attention patterns, which are already tuned for cross-attention selection.

\subsection{M.~1: Memory as Encoder-Input Prefix}
\label{sec:var1}

Persistent memory is compressed into $m$ soft tokens and prepended to the
encoder input, extending the prefix-tuning idea~\citep{li2021prefix} from
static task prompts to a dynamic, accumulating memory bank.
The encoder integrates memory and current input through
self-attention and produces a valid $Z \in \mathcal{M}_{E}$.  The decoder
remains entirely untouched.  Concretely,
\begin{align}
  S_t &= U_P P_{t-1} W_P
    \in \R^{m \times d},
    \label{eq:v1_proj} \\
  \tilde{x}_t &= \bigl[\;\underbrace{S_t}_{\text{$m$ soft tokens}};\;
                 x_t\;\bigr]
    \in \R^{(m+n) \times d},
    \label{eq:v1_input} \\
  \tilde{Z}_t &= E_{\freeze}(\tilde{x}_t), \qquad
  Z_t = \tilde{Z}_t[m{:}],
    \label{eq:v1_enc} \\
  \hat{y}_t &= D_{\freeze}(Z_t),
    \label{eq:v1_dec}
\end{align}
where $U_P \in \R^{m \times n_P}$ mixes the $n_P$ memory rows into $m$ prefix
slots and $W_P \in \R^{d \times d}$ is a learnable feature projection.  If
$n_P = m$, one may set $U_P = I$, reducing the prefix to
$S_t = P_{t-1} W_P$.

Memory is updated via an attention-coupled write rule that injects the
current turn's content into the memory bank:
\begin{align}
  Q &= Z_t W_Q,\quad K = P_{t-1} W_K,\quad V = Z_t W_V, \notag\\
  A_t &= \softmax\!\bigl(QK^\top\!/\!\sqrt{d}\bigr), \qquad
  P_t = \gamma P_{t-1} + A_t^\top V.
  \label{eq:v1_write}
\end{align}
$Q$ and $K$ perform content-based addressing between the current latent~$Z_t$
and the existing memory~$P_{t-1}$.  Crucially, values~$V$ are drawn from
$Z_t$, not from~$P$; $A_t^\top V$ aggregates the current turn's content into
$n_P$~memory rows, weighted by the addressing scores.  This ensures that new
information enters memory at every turn---without it, an all-zero
initialisation would remain zero indefinitely.

Figure~\ref{fig:v1} illustrates how memory is projected into a soft prefix
before the frozen encoder, with the current latent driving a write-back update.

\begin{center}
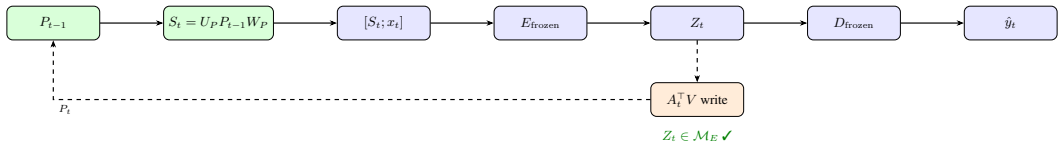

\begin{tikzpicture}[
  node distance=0.9cm and 1.5cm,
  block/.style={draw, rounded corners, minimum width=2.2cm,
                minimum height=0.8cm, align=center, font=\small},
  frozen/.style={block, fill=blue!10},
  mem/.style={block, fill=green!15},
  write/.style={block, fill=orange!15},
  arr/.style={-{Stealth[length=5pt]}, thick},
  darr/.style={-{Stealth[length=5pt]}, thick, dashed},
]
  \node[mem] (P) {$P_{t-1}$};
  \node[mem, right=of P] (proj) {$S_t = U_P P_{t-1} W_P$};
  \node[frozen, right=of proj] (cat) {$[S_t; x_t]$};
  \node[frozen, right=of cat] (enc) {$E_{\freeze}$};
  \node[frozen, right=of enc] (Z) {$Z_t$};
  \node[frozen, right=of Z] (dec) {$D_{\freeze}$};
  \node[frozen, right=of dec] (y) {$\hat{y}_t$};
  \node[write, below=1.0cm of Z] (write) {$A_t^\top V$ write};

  \draw[arr] (P) -- (proj);
  \draw[arr] (proj) -- (cat);
  \draw[arr] (cat) -- (enc);
  \draw[arr] (enc) -- (Z);
  \draw[arr] (Z) -- (dec);
  \draw[arr] (dec) -- (y);
  \draw[darr] (Z) -- (write);
  \draw[darr] (write) -| node[below right, font=\scriptsize]{$P_t$} (P);

  \node[below=0.2cm of write, text=green!50!black, font=\small\bfseries]
    {$Z_t \in \mathcal{M}_{E}$ \ding{51}};
\end{tikzpicture}

\captionof{figure}{M.~1 injects persistent memory as an encoder-input prefix
and writes the current latent back into memory through an attention-coupled
update.}
\label{fig:v1}
\end{center}

The trainable read-side parameters are $\{W_P\}$;
the write-side projections $\{W_Q, W_K, W_V\}$ and decay~$\gamma$ are frozen
(Sec.~\ref{sec:learning}).
Under Type~1 training, $\theta_{\mem}$ is optimised via
$\nabla_{\theta_{\mem}} \mathcal{L}$; under Type~2, $P_t$ accumulates at
inference with $\theta_{\mem}$ frozen.

\subsection{M.~2: Parallel Decoder Cross-Attention}
\label{sec:var2}

A parallel cross-attention layer is inserted in each decoder block to attend
to~$P$ independently of the frozen pathway, following the Flamingo
architecture~\citep{alayrac2022flamingo} which showed that interleaved
cross-attention can inject external information into a frozen LM.  The
original $Z$~route is untouched; memory influence is \emph{additive} via a
zero-initialised coefficient.  This method requires source-level access to
the decoder blocks of an open-weight model.  Here \emph{frozen} means that
the original decoder weights are not updated; it does not mean that
intermediate layers are inaccessible.

At decoder layer~$\ell$ with hidden state $s_t^{(\ell)}$:
\begin{align}
  c_{\mem}^{(\ell)} &= \mathrm{XAttn}_{\mem}^{(\ell)}
    \!\bigl(s_t^{(\ell)},\; P_{t-1}\bigr), \label{eq:v2_xattn} \\
  s_t^{(\ell)\prime} &= s_t^{(\ell)}
    + \mathrm{XAttn}_{\freeze}^{(\ell)}\!\bigl(s_t^{(\ell)},\, Z_t\bigr)
    + \underbrace{\beta^{(\ell)}}_{\text{init}=0}\, c_{\mem}^{(\ell)}.
    \label{eq:v2_merge}
\end{align}
At initialisation $\beta^{(\ell)}=0$, so the model falls back exactly to
the frozen baseline.  Memory is updated with the same attention-coupled
write rule as M.1:
\begin{equation}
  Q = Z_t W_Q,\; K = P_{t-1} W_K,\; V = Z_t W_V,\quad
  P_t = \gamma\, P_{t-1} + A_t^\top V,
  \label{eq:v2_write}
\end{equation}
where $A_t = \softmax(QK^\top/\sqrt{d})$ and $V$ is sourced from~$Z_t$
so that new content enters memory regardless of~$P$'s current state.
Figure~\ref{fig:v2} illustrates the frozen
cross-attention path running in parallel with the additive memory branch.

\begin{center}
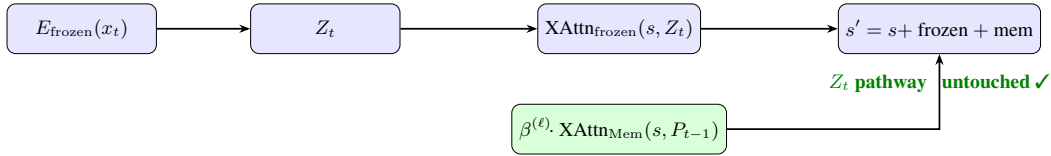

\begin{tikzpicture}[
  node distance=0.9cm and 1.5cm,
  block/.style={draw, rounded corners, minimum width=2.4cm,
                minimum height=0.8cm, align=center, font=\small},
  frozen/.style={block, fill=blue!10},
  mem/.style={block, fill=green!15},
  arr/.style={-{Stealth[length=5pt]}, thick},
]
  \node[frozen] (enc) {$E_{\freeze}(x_t)$};
  \node[frozen, right=of enc] (Z) {$Z_t$};
  \node[frozen, right=2.2cm of Z] (xattn) {XAttn$_{\freeze}(s, Z_t)$};
  \node[mem, below=0.8cm of xattn] (memx)
    {$\beta^{(\ell)}\!\cdot$ XAttn$_{\mem}(s, P_{t-1})$};
  \node[frozen, right=2.2cm of xattn] (sum) {$s' = s +$ frozen $+$ mem};

  \draw[arr] (enc) -- (Z);
  \draw[arr] (Z) -- (xattn);
  \draw[arr] (xattn) -- (sum);
  \draw[arr] (memx) -| (sum);

  \node[below=0.2cm of sum, text=green!50!black, font=\small\bfseries]
    {$Z_t$ pathway ~~untouched \ding{51}};
\end{tikzpicture}

\captionof{figure}{M.~2 preserves the frozen cross-attention route over the
current encoder latent and adds a parallel decoder memory branch scaled by a
learned coefficient.}
\label{fig:v2}
\end{center}

The per-layer parameters are
$\theta_{\mem} = \{W_Q^{\mem}, W_K^{\mem}, W_V^{\mem}, O^{\mem}, \beta^{(\ell)}\}$,
totalling ${\sim}16.8$M (0.6\%); the cross-attention projections are shared
across layers and $\beta^{(\ell)}$ adds one scalar per layer.

\smallskip\noindent\textbf{Implementation note.}
Because injecting into each frozen decoder block requires per-layer
hooks, our implementation approximates the per-layer read by computing
$\mathrm{XAttn}_{\mem}$ once using $Z_t$ as a proxy for internal decoder
states and blending with $\bar{\beta} = \mathrm{mean}(\beta^{(\ell)})$.
The result is passed as additional encoder positions, so the frozen
decoder's own per-layer projections provide implicit layer
specialisation.

\subsection{M.~3: Decoder KV Extension}
\label{sec:var3}

Persistent memory is projected into additional key--value pairs that are
concatenated alongside~$Z$ in the decoder's cross-attention, following the
KV-extension principle of Memorizing
Transformers~\citep{wu2022memorizing}, and leaving the original
$Z$~positions byte-for-byte preserved.
A shared zero-initialised projection maps $P$ into pseudo-encoder hidden
states; the frozen decoder then applies its own per-layer
$W_K^{(\ell)}$, $W_V^{(\ell)}$ to both the original $Z$ positions and the
memory extension:
\begin{align}
  H_{\mem} &= P_{t-1}\, W_{\mem}
    \in \R^{n_P \times d}, \label{eq:v3_proj} \\
  K^{(\ell)} &= W_K^{(\ell)}
    \bigl[\,Z_t \;;\; H_{\mem}\,\bigr]
    \in \R^{(n+n_P) \times d_k}, \label{eq:v3_K} \\
  V^{(\ell)} &= W_V^{(\ell)}
    \bigl[\,Z_t \;;\; H_{\mem}\,\bigr]
    \in \R^{(n+n_P) \times d_v}, \label{eq:v3_V} \\
  \hat{y}_t  &= D_{\freeze}\!\bigl(Q^{(\ell)},\, K^{(\ell)},\, V^{(\ell)}\bigr).
    \label{eq:v3_dec}
\end{align}
Queries $Q^{(\ell)}$ use the frozen $W_Q^{(\ell)}$.
Zero-initialised $W_{\mem}$ ensures no-regression at init; the frozen
per-layer projections provide implicit layer specialisation without extra
learned parameters.
The write rule is the same attention-coupled update as M.1
(Eq.~\eqref{eq:v1_write}):
\begin{equation}
  Q = Z_t W_Q,\quad K = P_{t-1} W_K,\quad V = Z_t W_V,\quad
  P_t = \gamma P_{t-1} + \softmax\!\bigl(QK^\top\!/\!\sqrt{d}\bigr)^\top V.
  \label{eq:v3_write}
\end{equation}
Figure~\ref{fig:v3} shows decoder keys and values extended with
memory-derived entries while the current encoder positions remain unchanged.

\begin{center}
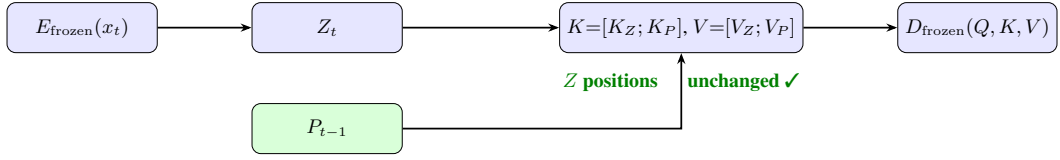

\begin{tikzpicture}[
  node distance=0.9cm and 1.5cm,
  block/.style={draw, rounded corners, minimum width=2.4cm,
                minimum height=0.8cm, align=center, font=\small},
  frozen/.style={block, fill=blue!10},
  mem/.style={block, fill=green!15},
  arr/.style={-{Stealth[length=5pt]}, thick},
]
  \node[frozen] (enc) {$E_{\freeze}(x_t)$};
  \node[frozen, right=of enc] (Z) {$Z_t$};
  \node[mem, below=0.8cm of Z] (P) {$P_{t-1}$};
  \node[frozen, right=2.5cm of Z] (kv)
    {$K{=}[K_Z; K_P]$, $V{=}[V_Z; V_P]$};
  \node[frozen, right=of kv] (dec) {$D_{\freeze}(Q, K, V)$};

  \draw[arr] (enc) -- (Z);
  \draw[arr] (Z) -- (kv);
  \draw[arr] (P) -| (kv);
  \draw[arr] (kv) -- (dec);

  \node[below=0.2cm of kv, text=green!50!black, font=\small\bfseries]
    {$Z$ positions ~~~~~unchanged \ding{51}};
\end{tikzpicture}

\captionof{figure}{M.~3 extends decoder keys and values with learned
projections of persistent memory while keeping the current encoder tokens on
their original path.}
\label{fig:v3}
\end{center}

The trainable read-side parameter is $W_{\mem} \in \R^{d \times d}$;
the write-side projections $\{W_Q, W_K, W_V\}$ and decay~$\gamma$ are frozen
(Sec.~\ref{sec:learning}).  Total added: ${\sim}4.2$M (0.1\%).

\subsection{M.~4: Hebbian / Associative Memory}
\label{sec:var4}

An outer-product Hebbian rule---the same write primitive used in linear
transformers and fast weight
programmers~\citep{schlag2021linear}---accumulates associative structure
in a matrix $M_t$, and the full read path is made explicit by injecting
the recalled memory through decoder KV extension.  This yields a
complete, experimentally realizable architecture rather than a generic
``inject somehow'' formulation.
Let $d_h$ denote the associative-memory dimension:
\begin{align}
  \tilde{M}_t &= \gamma M_{t-1}
      + \frac{1}{n}(Z_t W_{K,H})^\top (Z_t W_{V,H}), \notag \\
  M_t &= \tilde{M}_t \big/ \max\!\bigl(\lVert \tilde{M}_t\rVert_F,\,1\bigr)
      \;\in \R^{d_h \times d_h},
      \label{eq:v4_write} \\
  R_t &= (Z_t W_{Q,H}) M_{t-1}
      \in \R^{n \times d_h},
      \label{eq:v4_read} \\
  H_{\mem} &= R_t\, W_{\mem}
      \in \R^{n \times d},
      \label{eq:v4_proj} \\
  \hat{y}_t &= D_{\freeze}\!\Bigl(
      Q^{(\ell)},\;
      W_K^{(\ell)}\bigl[Z_t;\; H_{\mem}\bigr],\;
      W_V^{(\ell)}\bigl[Z_t;\; H_{\mem}\bigr]
    \Bigr).
      \label{eq:v4_dec}
\end{align}
Figure~\ref{fig:v4} shows the Hebbian matrix being read by the current latent
and exposed to the decoder as extra key-value memory.

\begin{center}
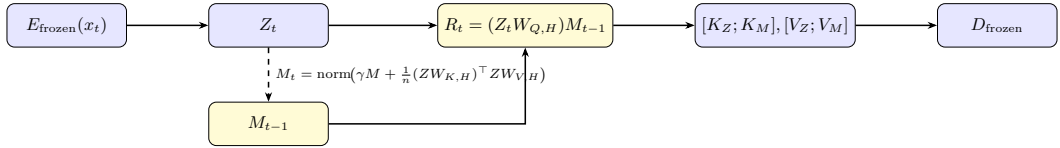

\begin{tikzpicture}[
  node distance=0.9cm and 1.5cm,
  block/.style={draw, rounded corners, minimum width=2.2cm,
                minimum height=0.8cm, align=center, font=\small},
  frozen/.style={block, fill=blue!10},
  mem/.style={block, fill=yellow!20},
  arr/.style={-{Stealth[length=5pt]}, thick},
  darr/.style={-{Stealth[length=5pt]}, thick, dashed},
]
  \node[frozen] (enc) {$E_{\freeze}(x_t)$};
  \node[frozen, right=of enc] (Z) {$Z_t$};
  \node[mem, below=1.0cm of Z] (M) {$M_{t-1}$};
  \node[mem, right=2.0cm of Z] (read) {$R_t = (Z_t W_{Q,H}) M_{t-1}$};
  \node[frozen, right=of read] (kv) {$[K_Z;K_M], [V_Z;V_M]$};
  \node[frozen, right=of kv] (dec) {$D_{\freeze}$};

  \draw[arr] (enc) -- (Z);
  \draw[arr] (Z) -- (read);
  \draw[arr] (M) -| (read);
  \draw[arr] (read) -- (kv);
  \draw[arr] (kv) -- (dec);
  \draw[darr] (Z) -- (M) node[midway, right, font=\scriptsize]
    {$M_t = \mathrm{norm}\!\bigl(\gamma M + \tfrac{1}{n}(ZW_{K,H})^\top ZW_{V,H}\bigr)$};
\end{tikzpicture}

\captionof{figure}{M.~4 stores associative structure in a Hebbian memory
matrix that is queried by the current latent and passed to the decoder as
additional memory.}
\label{fig:v4}
\end{center}

The trainable parameters are
$\theta_{\mem} = \{W_{Q,H}, W_{\mem}\}$;
the Hebbian write projections $\{W_{K,H}, W_{V,H}\}$ and decay~$\gamma$ are
frozen (Sec.~\ref{sec:learning}).
$W_{\mem}$ (the read projection from $d_h$ back to $d$) is zero-initialised so
that the memory branch is silent at startup (safe startup).
$W_{Q,H}$ uses small random initialisation, $\mathcal{N}(0, 0.02)$:
because $W_{\mem}$ is zero, the gradient signal for $W_{Q,H}$ comes from the
loss through $W_{\mem}$ once it becomes non-zero, and conversely the gradient
for $W_{\mem}$ requires $R_t \neq 0$, i.e.\ $W_{Q,H} \neq 0$.
Zero-initialising \emph{both} $W_{Q,H}$ and $W_{\mem}$ would create a
gradient deadlock.

\subsection{M.~5: Context-Gated Decoder Memory Branch}
\label{sec:var5}

Rather than modifying encoder outputs, a lightweight memory branch is inserted
\emph{inside the decoder}, using a content-dependent gate inspired by
Flamingo's tanh-gated cross-attention~\citep{alayrac2022flamingo}.
The branch reads from~$P$, and the gate controls how strongly its output
affects the decoder hidden state:
\begin{align}
  c_{\mem}^{(\ell)}
    &= \mathrm{XAttn}_{\mem}^{(\ell)}
       \!\bigl(s_t^{(\ell)},\, P_{t-1}\bigr),
       \label{eq:v5_read} \\
  g_t^{(\ell)}
    &= \sigma\bigl(W_g^{(\ell)} [s_t^{(\ell)}; c_{\mem}^{(\ell)}]
       + b_g^{(\ell)}\bigr),
       \label{eq:v5_gate} \\
  s_t^{(\ell)\prime}
    &= s_t^{(\ell)}
     + \mathrm{XAttn}_{\freeze}^{(\ell)}\!\bigl(s_t^{(\ell)}, Z_t\bigr)
     + g_t^{(\ell)} \odot c_{\mem}^{(\ell)}.
       \label{eq:v5_dec}
\end{align}
With $b_g^{(\ell)} < 0$ at initialisation, the memory branch starts nearly off,
so the model initially behaves like the frozen baseline and only later opens the
memory pathway where helpful.  Figure~\ref{fig:v5} illustrates the decoder-side
branch whose contribution is controlled by the learned context gate.

\begin{center}
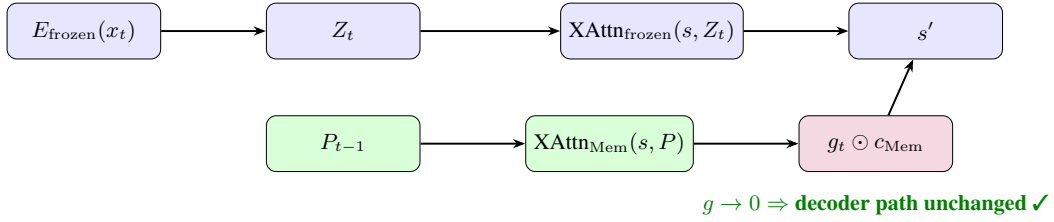

\begin{tikzpicture}[
  node distance=0.9cm and 1.5cm,
  block/.style={draw, rounded corners, minimum width=2.2cm,
                minimum height=0.8cm, align=center, font=\small},
  frozen/.style={block, fill=blue!10},
  mem/.style={block, fill=green!15},
  gate/.style={block, fill=purple!15},
  arr/.style={-{Stealth[length=5pt]}, thick},
]
  \node[frozen] (enc) {$E_{\freeze}(x_t)$};
  \node[frozen, right=of enc] (Z) {$Z_t$};
  \node[frozen, right=2.0cm of Z] (base) {XAttn$_{\freeze}(s,Z_t)$};
  \node[mem, below=0.8cm of Z] (P) {$P_{t-1}$};
  \node[mem, right=of P] (read) {XAttn$_{\mem}(s,P)$};
  \node[gate, right=of read] (gate) {$g_t \odot c_{\mem}$};
  \node[frozen, right=of base] (sum) {$s'$};

  \draw[arr] (enc) -- (Z);
  \draw[arr] (Z) -- (base);
  \draw[arr] (base) -- (sum);
  \draw[arr] (P) -- (read);
  \draw[arr] (read) -- (gate);
  \draw[arr] (gate) -- (sum);

  \node[below=0.2cm of gate, text=green!50!black, font=\small\bfseries]
    {$g \to 0 \Rightarrow$ decoder path unchanged \ding{51}};
\end{tikzpicture}

\captionof{figure}{M.~5 adds a memory read branch inside the decoder and
lets a learned gate decide when the auxiliary memory signal should influence
the frozen path.}
\label{fig:v5}
\end{center}

The trainable parameters are
$\theta_{\mem} = \{W_Q^{\mem}, W_K^{\mem}, W_V^{\mem}, W_g, b_g\}$;
the write-side projections $\{W_Q, W_K, W_V\}$ and decay~$\gamma$ are frozen
(Sec.~\ref{sec:learning}).

\smallskip\noindent\textbf{Implementation note.}
As with M.2, the per-layer decoder injection is approximated by computing
the gated memory read once with $Z_t$ as proxy for decoder hidden states
and passing the result as additional encoder positions.
The write rule is the same attention-coupled update as M.1
(Eq.~\eqref{eq:v1_write}):
\begin{equation}
  Q = Z_t W_Q,\quad K = P_{t-1} W_K,\quad V = Z_t W_V,\quad
  P_t = \gamma P_{t-1} + \softmax\!\bigl(QK^\top\!/\!\sqrt{d}\bigr)^\top V.
  \label{eq:v5_write}
\end{equation}

\subsection{M.~6: Slot-Based Memory with Sparse Write}
\label{sec:var6}

Memory is organised as $S$~fixed-size slots $P \in \R^{S \times d}$,
adopting the addressable-slot design of Neural Turing
Machines~\citep{graves2014neural}.  At each turn, only the top-$k$
addressed slots are updated; the read path is an explicit decoder KV
extension:
\begin{align}
  \bar{z}_t &= \frac{1}{n}\sum_{i=1}^{n} Z_t[i,:],
    \label{eq:v6_mean} \\
  a_t &= \softmax\!\bigl(\bar{z}_t W_a P_{t-1}^\top / \sqrt{d}\bigr)
    \in \R^{S},
    \label{eq:v6_addr} \\
  m_t[s] &= \mathbf{1}[s \in \topk(a_t, k)],
    \label{eq:v6_mask} \\
  u_t &= \bar{z}_t W_u \in \R^{d},
    \label{eq:v6_candidate} \\
  P_t[s] &= (1-m_t[s]) P_{t-1}[s]
    + m_t[s]\bigl(\gamma P_{t-1}[s] + (1-\gamma)u_t\bigr),
    \label{eq:v6_write} \\
  H_{\mem} &= P_t\, W_{\mem}, \label{eq:v6_proj} \\
  K^{(\ell)} &= W_K^{(\ell)}\bigl[Z_t;\; H_{\mem}\bigr],
  \quad
  V^{(\ell)} = W_V^{(\ell)}\bigl[Z_t;\; H_{\mem}\bigr].
    \label{eq:v6_kv}
\end{align}
Figure~\ref{fig:v6} shows sparse writes into a fixed slot bank that is later
read by the decoder as structured episodic memory.

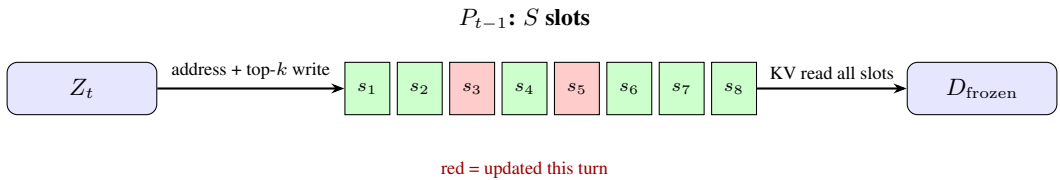
\begin{center}
\begin{tikzpicture}[
  node distance=0.8cm and 1.4cm,
  block/.style={draw, rounded corners, minimum width=2.0cm,
                minimum height=0.7cm, align=center, font=\small},
  frozen/.style={block, fill=blue!10},
  slot/.style={draw, minimum width=0.6cm, minimum height=0.7cm,
               font=\scriptsize, align=center},
  arr/.style={-{Stealth[length=5pt]}, thick},
]
  \node[frozen] (Z) {$Z_t$};
  \node[right=2.5cm of Z] (slots) {};

  \foreach \i/\c in {1/green!20, 2/green!20, 3/red!20, 4/green!20, 5/red!20,
                     6/green!20, 7/green!20, 8/green!20} {
    \node[slot, fill=\c, right=\i*0.7cm-0.7cm of slots.west] (s\i) {$s_\i$};
  }
  \node[above=0.3cm of s4, font=\small\bfseries] {$P_{t-1}$: $S$ slots};
  \node[frozen, right=2.0cm of s8] (dec) {$D_{\freeze}$};

  \draw[arr] (Z) -- (s1) node[midway, above, font=\scriptsize] {address + top-$k$ write};
  \draw[arr] (s8) -- (dec) node[midway, above, font=\scriptsize] {KV read all slots};

  \node[below=0.5cm of s4, font=\scriptsize, text=red!60!black]
    {red = updated this turn};
\end{tikzpicture}

\captionof{figure}{M.~6 writes to a fixed set of memory slots sparsely and
exposes the slot bank to the decoder as an explicit episodic memory.}
\label{fig:v6}
\end{center}

The trainable read-side parameter is $W_{\mem} \in \R^{d \times d}$;
the write-side projections $\{W_a, W_u\}$ and decay~$\gamma$ are frozen
(Sec.~\ref{sec:learning}).

\section{Training and Inference}
\label{sec:learning}

Each of the six methods introduces learnable parameters~$\theta_{\mem}$
trained while both encoder and decoder remain frozen.  Training
proceeds in two phases.

\paragraph{Gradient flow through frozen networks.}
``Frozen'' means that $E$ and~$D$ receive no weight updates---it does
\emph{not} mean that gradients cannot flow through them.  Both networks
act as fixed differentiable functions: in the backward pass the chain
rule propagates the loss gradient through the frozen decoder (and, for
M.~1, also through the frozen encoder) to reach~$\theta_{\mem}$.  This
is the same principle underlying prefix tuning, LoRA, and adapter
methods---the only parameters that receive gradient updates are those
belonging to the memory adapter.  The frozen backbone's role is to
provide both a fixed forward computation and a fixed gradient signal
that the adapter must learn to exploit.

In the first phase (\textbf{Type~1: supervised learning}), gradients flow
from the decoder loss through the read pathway and update $\theta_{\mem}$,
thereby learning \emph{how} to read and write memory effectively:
\begin{equation}
  \theta_{\mem} \leftarrow \theta_{\mem}
    - \eta \nabla_{\theta_{\mem}} \mathcal{L}
    \bigl(D_{\freeze}(\mathrm{Read}(Z_t, P)), \; y_t\bigr).
  \label{eq:type1}
\end{equation}

\paragraph{Write projections as fixed random maps.}
During Type~1 training the write rule
$P_t = \mathrm{Write}(P_{t-1}, Z_t)$ executes without gradients to
prevent the computation graph from growing across the full conversation
history.  Consequently, the write-side projections ($W_Q, W_K, W_V$ in most
methods) receive no gradient updates and remain at their random
initialisation throughout training.  These projections act as fixed random
maps that nonetheless preserve pairwise distances in the input
(Johnson--Lindenstrauss property), ensuring that distinct encoder outputs
produce distinguishable memory entries.  The read-side parameters
($W_P$, $W_{\mem}$, cross-attention heads, gates) are therefore the only
parameters that receive gradient updates; they learn to decode whatever
structure the fixed write rule deposits in~$P$.  Excluding the write-side
projections from the optimiser prevents unnecessary weight decay and avoids
misleading parameter counts.

In the second phase (\textbf{Type~2: conversational learning}),
$\theta_{\mem}$ is frozen but $P_t$ continues to accumulate.  Each
conversation enriches~$P$, improving the system's responses without any
gradient computation:
\begin{equation}
  P_t = \mathrm{Write}(P_{t-1}, Z_t)
  \quad \text{with $\theta_{\mem}$ fixed.}
  \label{eq:type2}
\end{equation}
This is the mechanism we call \emph{conversational learning}: the system
becomes more knowledgeable and personalised with every conversation, exactly
through repeated online updates while the frozen model weights remain fixed.
Figure~\ref{fig:learning-loops} contrasts the two phases.

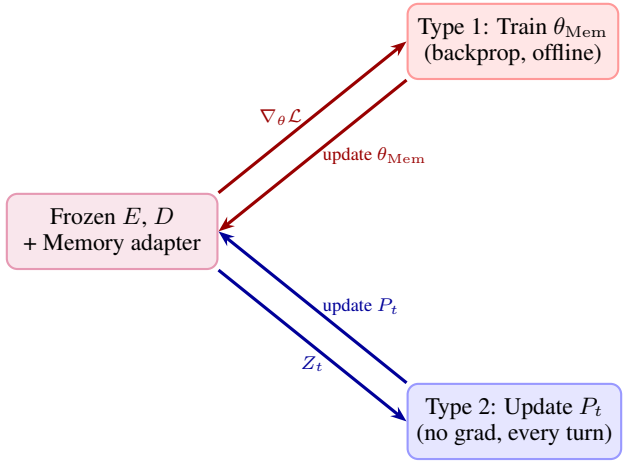
\begin{center}
\begin{tikzpicture}[
  node distance=1.2cm and 2.0cm,
  block/.style={draw, rounded corners, minimum width=2.8cm,
                minimum height=1.0cm, align=center, font=\small},
  type1/.style={block, fill=red!10, thick, draw=red!40},
  type2/.style={block, fill=blue!10, thick, draw=blue!40},
  both/.style={block, fill=purple!10, thick, draw=purple!40},
  arr/.style={-{Stealth[length=6pt]}, very thick},
  t1arr/.style={-{Stealth[length=6pt]}, very thick, red!60!black},
  t2arr/.style={-{Stealth[length=6pt]}, very thick, blue!60!black},
]
  \node[both] (model) {Frozen $E$, $D$\\+ Memory adapter};
  \node[type1, above right=1.5cm and 2.5cm of model] (t1)
    {Type~1: Train $\theta_{\mem}$\\(backprop, offline)};
  \node[type2, below right=1.5cm and 2.5cm of model] (t2)
    {Type~2: Update $P_t$\\(no grad, every turn)};

  \draw[t1arr] (model.north east) -- (t1.west)
    node[midway, left, font=\scriptsize, text=red!60!black]
    {$\nabla_\theta \mathcal{L}$};
  \draw[t1arr] (t1.south west) -- (model.east)
    node[midway, right, font=\scriptsize, text=red!60!black]
    {update $\theta_{\mem}$};
  \draw[t2arr] (model.south east) -- (t2.west)
    node[midway, below, font=\scriptsize, text=blue!60!black]
    {$Z_t$};
  \draw[t2arr] (t2.north west) -- (model.east)
    node[midway, right, font=\scriptsize, text=blue!60!black]
    {update $P_t$};
\end{tikzpicture}

\captionof{figure}{Two learning phases.  Type~1 updates memory
adapter parameters offline by backpropagation, whereas Type~2 updates the
persistent memory online at each turn with frozen model weights.}
\label{fig:learning-loops}
\end{center}

Table~\ref{tab:master} consolidates the six methods along key design
dimensions.

\begin{table}[!hbtp]
\centering
\caption{Six trained persistent-memory methods compared across key design
dimensions.  ``Primary-path safe'' means the original frozen
encoder--decoder route through $Z_t$ is preserved, even if an auxiliary
decoder-side memory branch is added.  ``Memory cost'' is per turn.}
\label{tab:master}
\renewcommand{\arraystretch}{1.15}
\resizebox{\textwidth}{!}{%
\begin{tabular}{clccccl}
\toprule
& \textbf{Method}
  & \textbf{Injection point}
  & \textbf{Primary-path safe}
  & \textbf{New params}
  & \textbf{Mem.\ cost}
  & \textbf{Write mechanism} \\
\midrule
1 & Encoder-input prefix
  & Before $E$ & \ding{51} & 4.2M & const.
  & $A_t^\top V$ \\
2 & Parallel decoder XAttn
  & Inside $D$ & \ding{51} & 16.8M & const.
  & $A_t^\top V$ \\
3 & Decoder KV extension
  & Inside $D$ xattn & \ding{51} & 4.2M & const.
  & $A_t^\top V$ \\
4 & Hebbian / associative
  & Decoder KV & \ding{51} & 1.0M & $O(d_h^2)$
  & Hebbian outer prod. \\
5 & Context-gated decoder branch
  & Inside $D$ & \ding{51} & 21.0M & const.
  & $A_t^\top V$ \\
6 & Slot-based sparse write
  & Decoder KV & \ding{51} & 4.2M & $O(Sd)$
  & Top-$k$ overwrite \\
\bottomrule
\end{tabular}}
\end{table}

\section{Evaluation}
\label{sec:evaluation}

The preceding sections specify \emph{how} each method reads and writes
persistent memory; this section specifies \emph{what} we measure and
\emph{how} we measure it.

\subsection{Forgetting-Curve Hypothesis}
\label{sec:knowledge_growth}

If persistent memory genuinely converts a stateless model into one that
\emph{retains conversation-specific information}, the effect should be
measurable as a function of how far in the past the supporting evidence was
written.  We therefore formalise evaluation around a \textbf{forgetting
curve} rather than a bag of unrelated scalar metrics.

For a question~$q$ asked after $T$ conversational turns, let $E_q$ denote the
set of supporting evidence turns and define the evidence lag by the oldest
required support:
\begin{equation}
  \ell_q = T - \min(E_q).
  \label{eq:lag_hypothesis}
\end{equation}
For each trained method~$m$, we compare two answers to the same question: one
with the learned persistent state intact, $\hat{y}^{\mathrm{mem}}_q(m)$, and
one from the same trained model with its persistent memory state forced to
zero, $\hat{y}^{0}_q(m)$.  The \emph{memory recall rate} is
\begin{equation}
  \rho_q(m) = \frac{\max\!\bigl(0,\;
    \operatorname{F1}(\hat{y}^{\mathrm{mem}}_q(m), y_q)
    - \operatorname{F1}(\hat{y}^{0}_q(m), y_q)\bigr)}
  {\max\!\bigl(1 - \operatorname{F1}(\hat{y}^{0}_q(m), y_q),\;\varepsilon\bigr)},
  \label{eq:retained_memory_hypothesis}
\end{equation}
where $y_q$ is the gold answer and $\varepsilon > 0$ is a small constant
that prevents division by zero.  The numerator is the F1 gain attributable
to persistent memory; the denominator is the headroom, i.e.\ the maximum
possible gain given the zero-memory baseline of that model.  The score is
therefore 100\% when memory brings the answer to perfect F1, and 0\% when
memory adds nothing.

For the \emph{stateless baseline}, there is no persistent state to ablate, so
the memory recall rate is identically zero for every question:
\begin{equation}
  \rho_q(\text{baseline}) = 0.
  \label{eq:baseline_zero_hypothesis}
\end{equation}
For an effective stateful system, the memory recall rate should be
largest at short lag and should gradually decrease as the relevant evidence
lies further in the past:
\begin{equation}
  \rho_q(m) \text{ should decrease as } \ell_q \text{ increases.}
  \label{eq:forgetting_hypothesis}
\end{equation}
The central experimental prediction is therefore a family of forgetting
curves: the baseline is flat at zero by construction, while stronger
trained persistent-memory methods start higher at short lag and decay more
slowly at long lag.  Without learned memory control, recall-rate curves
collapse toward zero in the stateless limit; the test is whether trained
adapter parameters~$\theta_{\mem}$ can produce large and durable positive
curves.

\subsection{Benchmarks and Protocol}
\label{sec:protocol}

\paragraph{Equal-input principle.}
To isolate the effect of persistent memory, \emph{every} condition---baseline
and all six memory methods---receives exactly the same encoder input~$x_t$ at
each turn: the current conversational turn only.  No method is given the full
history $[x_0, x_1, \ldots, x_t]$.  The baseline is therefore intentionally
\emph{short-sighted}: it encodes only the present turn and decodes an answer
with no access to prior sessions.  Memory methods receive the same~$x_t$ but
additionally condition the decoder on the persistent state~$P_{t-1}$
accumulated from all earlier turns.  Any non-zero memory recall rate is thus
attributable solely to the persistent state~$P$.

All conditions use the same released frozen encoder--decoder backbone, the same
tokenizer, and the same decoding rule.  Each example is processed turn by turn:
after every conversational turn the method-specific write rule updates~$P_t$;
when a question is asked the model answers using the current query and the
persistent memory state~$P_{t-1}$.  The stateless baseline uses the identical
backbone but no persistent memory.  Only
$\theta_{\mem}$ is optimised on the public training split with
teacher-forced answer loss; the encoder and decoder remain frozen throughout.
No method may use an external
summarizer, a different retriever, or a different pretrained model.

\paragraph{Implementation details.}
The frozen backbone is Flan-T5-XL (3B parameters) in bfloat16.
All memory adapters are trained with AdamW (learning rate $10^{-4}$,
weight decay $10^{-2}$, linear warmup of 200 steps, gradient norm clipped
at~1.0) for 10~epochs with batch size~2 and gradient accumulation~8
(effective batch~16).  Write-rule updates are detached from the
computation graph; truncated backpropagation through time uses a window
of $k{=}8$~turns.  Shared memory hyper-parameters are: bank size
$n_P{=}64$, write decay $\gamma{=}0.95$.  M.4 uses associative dimension
$d_h{=}256$; M.6 uses $S{=}64$ slots with top-$k{=}8$ writes per turn.
All experiments use a single seed (42) and a single NVIDIA GPU.

The primary benchmark is \textbf{LoCoMo}~\citep{maharana2024locomo}, a
long-term conversational-memory dataset with explicit QA supervision and
annotated evidence turns.  Those evidence annotations are the key ingredient
for the rebuilt evaluation: they let us place every question at a precise lag
in the past.  We do \emph{not} use MSC~\citep{xu2022msc} as a primary
score in this paper, because MSC lacks per-answer evidence locations and
therefore cannot support a clean forgetting curve.

\subsection{Forgetting-Curve Evaluation}
\label{sec:metrics}

We evaluate one quantity only: the \emph{forgetting curve}, expressed in
terms of the headroom-normalised \textbf{memory recall rate}.  For a
LoCoMo QA pair $q$ with gold answer $y_q$ and annotated evidence turns
$E_q$, let
\begin{equation}
  \ell_q = T - \min E_q
\end{equation}
denote the lag in turns from the oldest required fact to the end of the
conversation.

For each method $m$, we compute two answers to the same question:
\begin{align}
  \hat{y}^{\mathrm{mem}}_q(m) &= \text{answer from method $m$ after writing the conversation into } P, \\
  \hat{y}^{0}_q(m) &= \text{answer from the same trained method with its persistent state forced to zero.}
\end{align}
The memory recall rate of question $q$ is then
\begin{equation}
  \rho_q(m) = \frac{\max\!\bigl(0,\;
      F1(\hat{y}^{\mathrm{mem}}_q(m), y_q)
      - F1(\hat{y}^{0}_q(m), y_q)\bigr)}
  {\max\!\bigl(1 - F1(\hat{y}^{0}_q(m), y_q),\;\varepsilon\bigr)}.
  \label{eq:absolute_forgetting}
\end{equation}
The numerator is the F1 gain from persistent memory; the denominator is
the headroom---the maximum possible gain given the zero-memory
baseline.  The score therefore lies on a 0--100\% scale: 100\% means
memory brings the answer to perfect F1, 0\% means memory adds nothing.
This is \emph{not} baseline-relative.  It is a method-internal
quantity: how much of the available improvement room is filled by that
method's own memory.  The stateless baseline has no persistent state at all, so
\begin{equation}
  \rho_q(\text{M.0}) = 0 \qquad \text{for all questions } q.
\end{equation}

We bucket lags into five ranges,
$[0,32)$, $[32,64)$, $[64,128)$, $[128,256)$, and $[256,\infty)$ turns, average
$\rho_q(m)$ inside each bucket, and then fit a weighted non-increasing isotonic
curve across buckets.  The isotonic fit suppresses finite-sample noise while
preserving the forgetting prior that memory recall should not improve as
lag increases.  A stronger method therefore has a higher intercept at short lag
and a slower downward decay.

\subsection{Results}
\label{sec:results}

\begin{figure}[!hbtp]
\centering
\begin{tikzpicture}
\begin{axis}[
    name=ax1x,
    title={\textbf{1$\times$ capacity}},
    width=0.48\linewidth,
    height=6cm,
    xlabel={Evidence lag (turns)},
    ylabel={Memory recall rate (\%)},
    xmin=0.7, xmax=5.3,
    ymin=0, ymax=20,
    xtick={1,2,3,4,5},
    xticklabels={
      {$0$--$31$},{$32$--$63$},{$64$--$127$},{$128$--$255$},{$256+$}},
    x tick label style={rotate=45, anchor=east, font=\scriptsize},
    legend style={
      at={(1.02,0.5)},
      anchor=west,
      font=\tiny,
      cells={anchor=west},
    },
    grid=major,
    grid style={gray!30},
    cycle list name=color list,
    every axis plot/.append style={thick, mark=*,mark size=1.2pt},
    tick label style={font=\small},
    label style={font=\small},
]
\addplot coordinates {(1,0) (2,0) (3,0) (4,0) (5,0)};
\addlegendentry{Baseline}
\addplot coordinates {(1,0.0241) (2,0.0241) (3,0.0241) (4,0.0241) (5,0.0241)};
\addlegendentry{M.1 Prefix}
\addplot coordinates {(1,17.8482) (2,14.6477) (3,9.0221) (4,9.0221) (5,9.0221)};
\addlegendentry{M.2 XAttn}
\addplot coordinates {(1,0) (2,0) (3,0) (4,0) (5,0)};
\addlegendentry{M.3 KV Ext}
\addplot coordinates {(1,9.5079) (2,9.5079) (3,9.2264) (4,9.2264) (5,9.2264)};
\addlegendentry{M.4 Hebbian}
\addplot coordinates {(1,0.3571) (2,0.1038) (3,0.1038) (4,0.1038) (5,0.0873)};
\addlegendentry{M.5 Gated}
\addplot coordinates {(1,17.2109) (2,13.9087) (3,7.0810) (4,7.0810) (5,7.0810)};
\addlegendentry{M.6 Slot}
\end{axis}
\end{tikzpicture}%
\hfill
\begin{tikzpicture}
\begin{axis}[
    name=ax10x,
    title={\textbf{10$\times$ capacity}},
    width=0.48\linewidth,
    height=6cm,
    xlabel={Evidence lag (turns)},
    ylabel={},
    xmin=0.7, xmax=5.3,
    ymin=0, ymax=20,
    xtick={1,2,3,4,5},
    xticklabels={
      {$0$--$31$},{$32$--$63$},{$64$--$127$},{$128$--$255$},{$256+$}},
    x tick label style={rotate=45, anchor=east, font=\scriptsize},
    legend style={
      at={(1.02,0.5)},
      anchor=west,
      font=\tiny,
      cells={anchor=west},
    },
    grid=major,
    grid style={gray!30},
    cycle list name=color list,
    every axis plot/.append style={thick, mark=*,mark size=1.2pt},
    tick label style={font=\small},
    label style={font=\small},
]
\addplot coordinates {(1,0) (2,0) (3,0) (4,0) (5,0)};
\addlegendentry{Baseline}
\addplot coordinates {(1,10.7489) (2,10.7489) (3,9.3011) (4,9.3011) (5,9.1986)};
\addlegendentry{M.1 Prefix}
\addplot coordinates {(1,11.9034) (2,11.9034) (3,9.8790) (4,9.8790) (5,9.8790)};
\addlegendentry{M.2 XAttn}
\addplot coordinates {(1,15.5796) (2,15.5796) (3,9.6895) (4,9.6895) (5,9.6895)};
\addlegendentry{M.3 KV Ext}
\addplot coordinates {(1,15.8648) (2,11.1905) (3,10.3205) (4,10.3205) (5,10.3205)};
\addlegendentry{M.4 Hebbian}
\addplot coordinates {(1,11.2245) (2,7.6223) (3,7.6223) (4,7.6223) (5,7.6223)};
\addlegendentry{M.5 Gated}
\addplot coordinates {(1,13.8549) (2,10.6037) (3,10.6037) (4,10.2078) (5,9.6614)};
\addlegendentry{M.6 Slot}
\end{axis}
\end{tikzpicture}
\caption{%
  Forgetting curves on LoCoMo at two memory-capacity scales.  Each
  point is the memory recall rate from
  Eq.~\eqref{eq:absolute_forgetting}, smoothed by a weighted
  non-increasing isotonic fit.  \textbf{Left}
  (1$\times$): three methods (M.1, M.3, M.5) collapse; M.2~XAttn
  and M.6~Slot dominate.  \textbf{Right}
  (10$\times$): all six methods produce non-trivial curves;
  M.4~Hebbian is strongest at long lag.  Higher and flatter
  curves indicate stronger memory.%
}
\label{fig:forgetting_curve}
\end{figure}
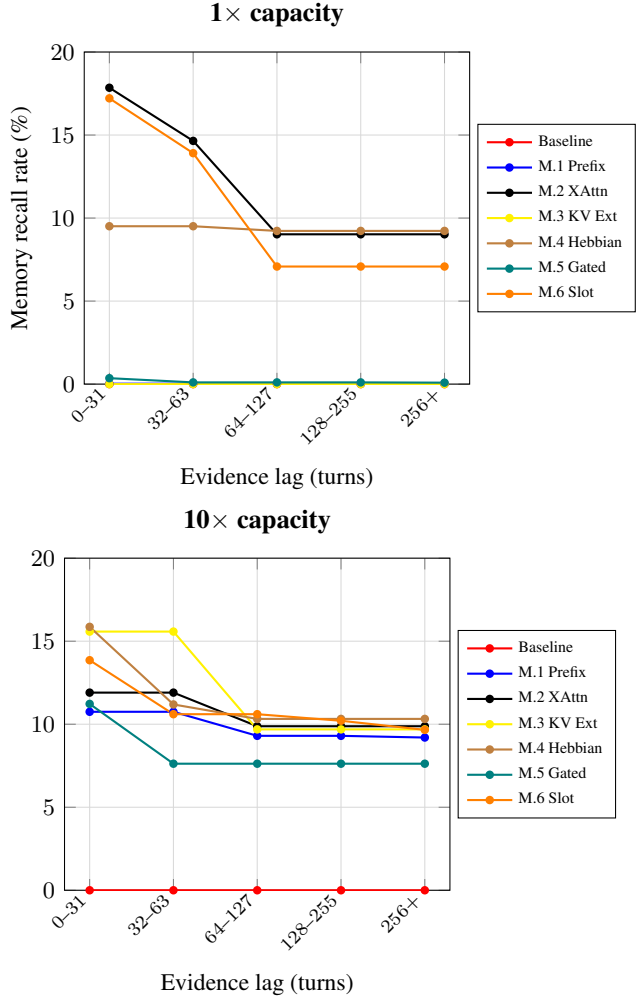

Figure~\ref{fig:forgetting_curve} is the primary empirical result.  We
evaluate two memory-capacity scales: \textbf{1$\times$} ($n_p\!=\!64$,
$d_h\!=\!256$, $n_{\text{slots}}\!=\!64$) and \textbf{10$\times$}
($n_p\!=\!640$, $d_h\!=\!810$, $n_{\text{slots}}\!=\!640$).  The
short-lag height measures write effectiveness: how much
conversation-specific information is available immediately after
storage.  The slope measures resistance to overwrite and interference:
how slowly that information decays as the relevant evidence recedes
into the past.

\begin{table}[!hbtp]
\centering
\caption{Memory recall rate (\%) by lag bucket on LoCoMo (smoothed,
non-increasing isotonic fit) at two capacity scales.  The baseline is
identically zero by construction.  $n$ gives the number of QA pairs
per bucket.  Higher $\to$ stronger memory retention.}
\label{tab:results}
\small
\begin{tabular}{@{}l ccccc c@{}}
\toprule
& \multicolumn{5}{c}{\textbf{Lag bucket (turns)}} & \\
\cmidrule(lr){2-6}
\textbf{Method} & 0--31 & 32--63 & 64--127 & 128--255 & 256+ & \textbf{Mean} \\
\midrule
$n$ (samples) & 28 & 24 & 62 & 130 & 395 & \\
\midrule
\multicolumn{7}{@{}l}{\textit{1$\times$ capacity} ($n_p\!=\!64$, $d_h\!=\!256$)} \\[2pt]
M.0 Baseline   & 0.00 & 0.00 & 0.00 & 0.00 & 0.00 & 0.00 \\
M.1 Prefix     & 0.02 & 0.02 & 0.02 & 0.02 & 0.02 & 0.02 \\
M.3 KV Ext     & 0.00 & 0.00 & 0.00 & 0.00 & 0.00 & 0.00 \\
M.5 Gated      & 0.36 & 0.10 & 0.10 & 0.10 & 0.09 & 0.15 \\
M.4 Hebbian    & 9.51 & 9.51 & 9.23 & 9.23 & 9.23 & 9.34 \\
M.6 Slot       & 17.21 & 13.91 & 7.08 & 7.08 & 7.08 & 10.47 \\
M.2 XAttn      & 17.85 & 14.65 & 9.02 & 9.02 & 9.02 & 11.91 \\
\midrule
\multicolumn{7}{@{}l}{\textit{10$\times$ capacity} ($n_p\!=\!640$, $d_h\!=\!810$)} \\[2pt]
M.0 Baseline   & 0.00 & 0.00 & 0.00 & 0.00 & 0.00 & 0.00 \\
M.5 Gated      & 11.22 & 7.62 & 7.62 & 7.62 & 7.62 & 8.34 \\
M.1 Prefix     & 10.75 & 10.75 & 9.30 & 9.30 & 9.20 & 9.86 \\
M.2 XAttn      & 11.90 & 11.90 & 9.88 & 9.88 & 9.88 & 10.69 \\
M.3 KV Ext     & 15.58 & 15.58 & 9.69 & 9.69 & 9.69 & 12.05 \\
M.6 Slot       & 13.85 & 10.60 & 10.60 & 10.21 & 9.66 & 10.99 \\
M.4 Hebbian    & 15.86 & 11.19 & 10.32 & 10.32 & 10.32 & 11.60 \\
\bottomrule
\end{tabular}
\end{table}

\paragraph{Interpretation.}
The evaluation isolates persistent memory itself rather than generic
question-answering ability.  A method receives credit only for answer quality
that vanishes when its own persistent state is ablated, normalised by
the headroom available to that method.  The baseline is
therefore exactly flat at zero, while trained memory methods show positive
short-lag recall followed by gradual decay.
Because the score is headroom-normalised, it sits on a 0--100\% scale
and directly measures what fraction of the remaining improvement room
the persistent memory fills.

The figure separates two properties that were conflated by the previous
multi-metric protocol.  Short-lag height measures whether a method writes
useful content into memory at all.  Long-lag decay measures whether that
content survives overwriting and interference.  A stronger architecture is one
whose curve stays higher for longer.

\paragraph{1$\times$ capacity.}
At the smaller scale (Figure~\ref{fig:forgetting_curve}, left;
Table~\ref{tab:results}, upper block), the methods separate into two
tiers.  M.2~XAttn and M.6~Slot dominate with short-lag recall above
17\% and long-lag scores around 9\% and 7\%, respectively.
M.4~Hebbian is the most stable: its curve is nearly flat at
${\sim}9.3\%$ across all buckets, indicating strong resistance to
overwrite.  In contrast, M.1~Prefix, M.3~KV~Ext, and M.5~Gated
collapse to near-zero---their small memory banks cannot sustain useful
state.  The 1$\times$ ordering is therefore
M.2~XAttn $>$ M.6~Slot $>$ M.4~Hebbian $\gg$ M.5~Gated $>$
M.1~Prefix $\approx$ M.3~KV~Ext $\approx$ Baseline.

\paragraph{10$\times$ capacity.}
At the larger scale (Figure~\ref{fig:forgetting_curve}, right;
Table~\ref{tab:results}, lower block), \emph{all six methods produce
non-trivial forgetting curves}.  M.4~Hebbian now leads with the
highest long-lag score (10.3\%) and the best mean (11.6\%).
M.3~KV~Ext, which was dead at 1$\times$, achieves the highest
short-lag recall (15.6\%) and a strong mean (12.0\%).  M.6~Slot
remains consistently strong across all buckets.  M.5~Gated, which
collapsed at 1$\times$, now reaches 11.2\% short-lag recall.

The capacity effect is the most striking result: three methods
(M.1, M.3, M.5) fail completely at 1$\times$ but succeed at
10$\times$, demonstrating that memory bank size is a critical
hyperparameter---not just a scaling convenience.  The methods that
succeed at both scales (M.2, M.4, M.6) employ write mechanisms
that are inherently more selective: attention-coupled writes,
associative updates, or sparse top-$k$ slot addressing.

The fact that all six trained adapters produce non-trivial
memory-recall curves at sufficient capacity answers the secondary
question posed in the introduction: the frozen decoder's
cross-attention does possess sufficient representational slack to
attend usefully to memory entries projected by a trained
adapter---the bottleneck is the quality of the write and read
pathway and the capacity of the memory bank.

\subsection{Cumulative Knowledge Curve}
\label{sec:knowledge_accumulation}

The forgetting curve measures how well memory \emph{resists decay} after
information is written.  A complementary question is whether persistent
memory enables \textbf{knowledge accumulation}: does the model know
progressively more as additional sessions are processed?

We formalise this as a \emph{cumulative knowledge curve}.  Let
$\mathcal{S}_1, \mathcal{S}_2, \ldots, \mathcal{S}_n$ denote the sessions
in a LoCoMo conversation.  After writing all turns in sessions
$\mathcal{S}_1$ through~$\mathcal{S}_s$, we probe every QA pair whose
annotated evidence sessions are fully contained within the processed
sessions, i.e.\ $E_q^{\mathrm{sess}} \subseteq \{1, \ldots, s\}$.  The
session-level knowledge score is
\begin{equation}
  K_s = \frac{1}{|Q_{\leq s}|} \sum_{q \in Q_{\leq s}}
        \operatorname{F1}\!\bigl(\hat{y}_q^{\mathrm{mem}}, y_q\bigr),
  \label{eq:knowledge_curve}
\end{equation}
where $Q_{\leq s}$ is the set of answerable questions at session~$s$.
The net knowledge gain is summarised by
\begin{equation}
  \Delta K = K_n - K_1,
  \label{eq:delta_k}
\end{equation}
which is positive when the memory accumulates knowledge and zero (or
negative) when it merely decays.  For the stateless baseline, every
probed answer is produced without access to prior context, so $K_s$ is
approximately constant across sessions.  A rising $K_s$ curve therefore
provides direct visual evidence of knowledge growth that is impossible
in a memoryless system.

\paragraph{Knowledge-accumulation results (1$\times$ capacity).}
Table~\ref{tab:knowledge} reports the terminal knowledge
$K_{30}$ and net gain $\Delta K$ for each method.  The stateless
baseline accumulates a surprising $\Delta K = 5.6\%$---a ceiling
effect from the frozen encoder's raw QA ability improving as more
context is injected into the current prompt.  Among memory methods,
M.6~Slot achieves the highest $\Delta K = 9.7\%$ and the highest
terminal $K_{30} = 9.7\%$, followed by M.4~Hebbian ($\Delta K =
7.8\%$) and M.2~XAttn ($\Delta K = 7.3\%$).  These are the same
three methods that succeed on the forgetting curve at 1$\times$
capacity, confirming that write quality determines both retention
and accumulation.  M.1~Prefix, M.3~KV~Ext, and M.5~Gated show
near-zero knowledge growth ($\Delta K < 0.2\%$), consistent with
their collapsed forgetting curves.

\begin{table}[!hbtp]
\centering
\caption{Knowledge accumulation on LoCoMo (1$\times$ capacity).
$K_{30}$ is the terminal knowledge score after all 30 sessions;
$\Delta K = K_{30} - K_1$ measures net knowledge gain.
Higher values indicate stronger persistent memory.}
\label{tab:knowledge}
\small
\begin{tabular}{@{}l cc@{}}
\toprule
\textbf{Method} & $K_{30}$ (\%) & $\Delta K$ (\%) \\
\midrule
M.0 Baseline   & 5.57 & 5.57 \\
M.1 Prefix     & 0.00 & 0.00 \\
M.3 KV Ext     & 0.00 & 0.00 \\
M.5 Gated      & 0.17 & 0.17 \\
M.2 XAttn      & 11.04 & 7.34 \\
M.4 Hebbian    & 10.62 & 7.84 \\
M.6 Slot       & 9.71 & 9.71 \\
\bottomrule
\end{tabular}
\end{table}

\section{Discussion}
\label{sec:discussion}

\subsection{Why training is necessary}
\label{sec:zero_train_fail}

If persistent memory is built entirely from frozen encoder outputs and
exposed to the decoder without learned projections, the decoder receives
states that were never optimised for selective long-range retrieval.
The failure is structural: as more history is concatenated or cached,
useful entries compete with irrelevant ones inside the same softmax,
attention mass disperses, and the contribution of any single remembered
fact shrinks.  The frozen encoder manifold~$\mathcal{M}_E$ matches the
pre-trained cross-attention interface, but it provides no mechanism for
deciding what to retain, compress, or foreground across long lags.

The forgetting-curve results confirm this analysis.  At 1$\times$
capacity, methods whose projections are closer to the raw encoder
output (M.1~Prefix, M.3~KV~Ext) collapse entirely, while methods
with learned selective writes (M.2~XAttn, M.4~Hebbian, M.6~Slot)
achieve recall rates above 9\%.  At 10$\times$ capacity, even the
weaker methods recover, but the gap remains: methods with richer
write rules consistently outperform simpler ones.  Training is
therefore necessary---and so is sufficient memory capacity: the
adapter must learn to map persistent state back into a representation
the frozen decoder can use, and the bank must be large enough to
store it.

\subsection{Adapter interference}
\label{sec:interference}

A trained memory adapter modifies the decoder's cross-attention pathway.
Even when the persistent memory is empty—immediately after a reset, or
before any conversation has begun—the adapter's projections inject into
the decoder and may displace the pre-trained knowledge that the frozen
backbone already possesses.  We quantify this risk with two complementary
metrics.

\paragraph{Adapter tax.}
Let $\widetilde{F}_{\text{base}}$ be the raw token-F1 of the stateless
baseline (no adapter, no memory) and let
$\widetilde{F}^{0}_{m}$ be the F1 of method~$m$ with its adapter
attached but its memory state forced to zero.  The \emph{adapter tax} is
\begin{equation}
  \text{Tax}_{m} = \widetilde{F}_{\text{base}} - \widetilde{F}^{0}_{m}.
  \label{eq:adapter_tax}
\end{equation}
Positive values mean that the adapter degrades the model below
its unmodified stateless performance, negative values mean the adapter
happens to help even without any stored memory.

\paragraph{Net benefit.}
The quantity reviewers ultimately care about is whether adding persistent
memory gives a net improvement over the original model:
\begin{equation}
  \text{Benefit}_{m} = \widetilde{F}^{\mathrm{mem}}_{m} - \widetilde{F}_{\text{base}}.
  \label{eq:net_benefit}
\end{equation}
A method is worthwhile if and only if $\text{Benefit}_{m} > 0$, i.e.\ the
memory's contribution outweighs any adapter interference.

\paragraph{Why interference should be small.}
Because the backbone is 100\% frozen, the adapter's only effect on the
decoder is through the additional cross-attention entries it provides.
When the memory bank is zeroed, these entries carry near-zero magnitude;
a well-trained softmax distributes negligible attention mass to them,
leaving the decoder's original computation approximately intact.  Methods
with explicit gating (M.5, M.6) can learn to suppress the memory pathway
entirely when memory is uninformative.  The adapter tax is therefore
expected to be small, and the net benefit should track the forgetting-curve
results closely.

Both metrics are computed from the same $\text{F1}_{\text{mem}}$ and
$\text{F1}_{\text{zero}}$ values that the forgetting-curve evaluation
already records, so no additional experiments are required.
Table~\ref{tab:interference} reports the mean adapter tax, net benefit,
and raw F1 scores across lag buckets.

\begin{table}[t]
\centering
\caption{Adapter interference analysis.
  Tax $>0$ means the adapter degrades baseline knowledge when memory is
  empty; Benefit $>0$ means memory helps more than the adapter hurts.
  Baseline mean F1 = 6.44\% at both scales (stateless, independent of
  adapter capacity).}
\label{tab:interference}
\small
\begin{tabular}{@{}l rr rr@{}}
\toprule
& \multicolumn{2}{c}{\textbf{1$\times$ capacity}}
& \multicolumn{2}{c}{\textbf{10$\times$ capacity}} \\
\cmidrule(lr){2-3}\cmidrule(lr){4-5}
Method & Tax (\%) & Benefit (\%) & Tax (\%) & Benefit (\%) \\
\midrule
M.1 Prefix   & +2.38 &  $-$6.42 & +4.23 & +4.00 \\
M.2 XAttn    & +3.39 &  +6.76   & +3.39 & +5.09 \\
M.3 KV Ext   & +2.38 &  $-$6.44 & +4.23 & +6.26 \\
M.4 Hebbian  & +3.39 &  +3.10   & +3.39 & +5.46 \\
M.5 Gated    & +3.39 &  $-$6.30 & +3.39 & +1.83 \\
M.6 Slot     & +2.38 &  +5.75   & +4.23 & +5.22 \\
\bottomrule
\end{tabular}
\end{table}

\paragraph{Empirical observations.}
At 1$\times$ capacity, three methods (M.1~Prefix, M.3~KV~Ext,
M.5~Gated) show \emph{negative} net benefit---the adapter hurts more
than the memory helps, consistent with the capacity collapse observed in
the forgetting curves.  In contrast, M.2~XAttn (+6.76\%), M.6~Slot
(+5.75\%), and M.4~Hebbian (+3.10\%) produce positive net benefit even
at low capacity.  At 10$\times$ capacity, \emph{all six} methods yield
positive net benefit (range +1.8--6.3\%), confirming that sufficient
memory capacity overcomes the adapter tax.  The tax itself is modest
(2--4\% across all conditions), validating the theoretical prediction
that frozen-backbone adapters introduce only minor interference.

\subsection{Limitations and scope}

This work is an intentionally constrained pilot study.  All six methods
are instantiated on a single frozen encoder--decoder backbone
(Flan-T5-XL, 3B parameters) with a single evaluation dataset (LoCoMo)
and minimal compute.  Whether the same architectural principles transfer
to decoder-only, encoder-only, or other-scale models must be validated
separately.  Absolute recall rates remain modest (up to $\approx$12\%),
which is expected given that 100\% of backbone weights are frozen and
the adapter budget is small.

Critically, these limitations are \emph{by design}: the purpose of the
pilot is to demonstrate feasibility under worst-case resource constraints,
not to optimise absolute performance.  We expect that relaxing any of
the following constraints will yield substantially stronger results:
\begin{enumerate}[nosep]
  \item \textbf{Unfreezing the backbone.}  End-to-end training would
    allow the encoder to learn \emph{what} to write and the decoder to
    learn \emph{how} to read persistent memory, rather than forcing
    the adapter alone to bridge both gaps.
  \item \textbf{Larger models.}  Bigger LLMs (e.g.\ 70B+ decoder-only)
    possess richer internal representations; persistent memory injected
    into these representations should carry more information per slot.
  \item \textbf{Larger and more diverse data.}  Training on corpora
    beyond a single multi-session dialogue benchmark will improve
    generalisation of the write and read operations.
  \item \textbf{Larger memory banks.}  Our 10$\times$ scale uses
    $n_P=640$ slots.  Because the memory bank is a numerical array
    decoupled from the backbone, it can be scaled by orders of
    magnitude---potentially millions of slots---with no change to
    per-turn inference cost.
\end{enumerate}
Pursuing these directions requires industrial-scale compute and is
beyond the scope of this study, but the design-space taxonomy and
evaluation protocol established here provide the foundation for
such work.

\subsection{Broader implications}

\paragraph{End-to-end training and conversational learning.}
This pilot study trains only a minimal memory adapter ($\theta_{\mem}$)
while the entire backbone remains frozen.  This is the most
resource-constrained setting possible, and it already demonstrates
non-trivial memory recall.  The full potential of persistent
latent-space memory would be realised when the \emph{entire LLM is
trained end-to-end} with its memory bank, learning simultaneously what
to store and how to use it.  An industrial-scale effort---training a
70B+ model on diverse multi-session corpora with a persistent memory
bank millions of slots large---would couple the backbone's representation
power with the memory's persistence, a combination that our frozen
setup deliberately excludes.  We expect such a system to outperform our
pilot results by a wide margin; our contribution is to show that the
underlying mechanism is sound and to map the design space that
large-scale training should explore.

More broadly, persistent memory opens the door to \emph{conversational
learning}: every interaction updates the bank, and the model becomes
more informed with each turn, driven by ordinary dialogue rather than
curated datasets or reward signals.  Existing LLMs can be
\emph{retrofitted} by installing a memory adapter and retraining---the
backbone architecture need not change.

\paragraph{Scalability.}
A latent memory bank is a compact numerical array whose capacity can
grow without increasing the per-turn inference cost of the backbone.
Our experiments use at most $n_P=640$ slots ($\approx$5\,MB at
\texttt{float32}); in principle the bank can scale to \emph{millions}
of slots at modest storage cost, far exceeding the lifetime of any
human conversation.  Unlike text-level memory systems that must
retokenise growing text stores---incurring cost proportional to memory
size---latent memory is read through a fixed-dimension attention
operation, so per-turn inference cost is independent of how much
history has been stored.

\paragraph{Latent-space memory as a cognitive substrate.}
Biological brains do not retain verbatim transcripts; they maintain
distributed, continuously updated representations that support
recognition, abstraction, and generalisation.  Persistent latent-space
memory mirrors this organisation more closely than text-level retrieval,
and connects naturally to neuroscience-inspired architectures such as
complementary learning systems~\citep{kumaran2016cls} and
attention-coupled lateralised memory~\citep{jeong2026lateral}.
Because LLMs already represent knowledge as continuous activations, a
persistent memory that operates in the same latent space is a more
natural substrate for core cognitive operations---reading, updating,
generalisation, and compositional extension---than a symbolic or
textual buffer.

\section{Conclusion}
\label{sec:conclusion}

This paper presents a proof-of-concept pilot study: persistent memory
that lives entirely in the \emph{latent space} of a frozen
encoder--decoder LLM (Flan-T5-XL, 3B parameters), with only small
trainable memory adapters and minimal compute.  Under these deliberately
severe constraints, we show that the idea \emph{works}: six adapter
architectures spanning three injection points and four write mechanisms
all produce non-trivial memory-recall curves at 10$\times$ capacity,
while three collapse at 1$\times$---revealing capacity as a critical
design parameter.  M.2~XAttn and M.6~Slot dominate at low capacity;
M.4~Hebbian leads at high capacity.  The cumulative knowledge curve
confirms that the strongest methods accumulate knowledge steadily over
30~sessions ($\Delta K$ up to 9.7\%), while collapsed methods show no
growth.

The broader implication is that latent-space persistent memory is not
one mechanism but a \emph{design space}---one whose dimensions (write
rule, read path, capacity) have measurable consequences that would be
invisible in text-level memory systems.  Our pilot maps this design
space under worst-case conditions; the natural next step is
industrial-scale exploration: end-to-end training of large LLMs
(70B+) with memory banks scaled to millions of slots and diverse
multi-session corpora.  Because the memory bank is a compact numerical
array decoupled from the backbone, existing pre-trained models can be
retrofitted with persistent memory by installing an adapter and
retraining---no architectural redesign is required.  We believe that
full-scale training will show dramatically stronger results; this study
establishes the feasibility, the taxonomy, and the evaluation protocol
that such efforts need.

\section*{Acknowledgements}
The author thanks Inha University in Tashkent for research support.
This work reflects the author's ongoing inquiry into nature and human cognition.

\end{document}